\documentclass[preprint,12pt,numbers,sort&compress]{elsarticle}

\usepackage{ifthen}

\usepackage{amsmath}
\usepackage{amssymb,bm,bbm,amstext}
\usepackage{dsfont}
\usepackage[latin1]{inputenc}

\usepackage{ifthen}

\usepackage{color}   
\usepackage[vlined,linesnumbered,ruled,titlenotnumbered]{algorithm2e}

\usepackage{url}

\usepackage{multirow,rotating}
\usepackage{pdflscape}

\usepackage{balance}
\usepackage{multirow}

\usepackage{siunitx} 

\usepackage{xstring} 

\usepackage{color,graphicx}

\usepackage[labelfont=bf,font=footnotesize,indention=0.0cm,margin=0cm]{caption}  

\usepackage{tikz}
\usetikzlibrary{shapes,decorations.shapes}
\usepackage{pgfplots}
\usepackage{pgfplotstable}

\usepackage{pdfpages}

\newcommand{\plotAGE}{\includegraphics[height=2.5mm]{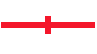}}
\newcommand{\plotNSGATWO}{\includegraphics[height=2.5mm]{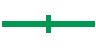}}
\newcommand{\plotSPEATWO}{\includegraphics[height=2.5mm]{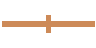}}
\newcommand{\plotIBEA}{\includegraphics[height=2.5mm]{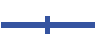}}
\newcommand{\plotSMSEMOA}{\includegraphics[height=2.5mm]{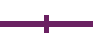}}

\newcommand{\R}{\mathbb{R}}

\newcommand{\ignore}[1]{}

\newtheorem{definition}{Definition}

\usepackage{xspace}
\newcommand{\CAC}{\textsc{CornersAndCentre}\xspace}
\newcommand{\LC}{\textsc{LinearCombinations}\xspace}
\newcommand{\NS}{\textsc{NoSeed}\xspace}

\begin{document}
\begin{frontmatter}

\title{Seeding the Initial Population of Multi-Objective Evolutionary Algorithms: A Computational Study}
\author[FSU]
{Tobias Friedrich}
\author[ADL]{Markus Wagner*}\corref{cor2}

\cortext[ADL]{Corresponding author. Phone: +61 8 8313 5405, fax: +61 8 8313 4366,\newline
email: markus.wagner@adelaide.edu.au.}
\address[FSU]{Friedrich-Schiller-Universit\"at Jena, Germany}
\address[ADL]{University of Adelaide, Australia}

\begin{abstract}
Most experimental studies initialize the population of evolutionary algorithms with random genotypes. In practice, however, optimizers are typically seeded with good candidate solutions either previously known or created according to some problem-specific method. This \emph{seeding} has been studied extensively for single-objective problems. For multi-objective problems, however, very little literature is available on the approaches to seeding and their individual benefits and disadvantages. 
In this article, we are trying to narrow this gap via a comprehensive computational study 
on common real-valued test functions. We investigate the effect of two seeding techniques for five algorithms on 48~optimization problems with 2, 3, 4, 6, and 8 objectives. We observe that some functions (e.g.,~DTLZ4 and the LZ family) benefit significantly from seeding, while others (e.g.,~WFG) profit less. The advantage of seeding also depends on the examined algorithm.
\end{abstract}


\end{frontmatter}

\sloppy

\section{Introduction}

In many real-world applications trade-offs between conflicting objectives
play a crucial role. As an example, consider route planning, where one objective might be
travel time and another fuel consumption. For such problems, we need
specialized optimizers that determine the Pareto front of mutually non-dominated solutions.
There are several established evolutionary multi-objective evolutionary algorithms (MOEA) and
many comparisons on various test functions. However, most of them start with
random initial solutions. 
In practice, however, a good initial \emph{seeding} can make problem solving approaches competitive that would otherwise be inferior.

If prior knowledge exists or can be
generated at a low computational cost, good initial estimates may generate
better solutions with faster convergence.
For single-objective evolutionary algorithms, methods such as seeding have been studied for about two decades; see, e.g., \cite{grefenstette1987incorporating, harik2000linkage,hopper2001empirical,liaw2000hybrid,yang2002hybrid} for studies and examples (see~\cite{Kazimipour2014seedingEA} for a recent categorization). For example, the effects of seeding for the Traveling Salesman Problem (TSP) and the job-shop scheduling problem (JSSP) were investigated in~\cite{oman2001}. 
The algorithms were seeded with known good solutions in the initial
population, and it was found that the results were significantly improved on the TSP but not
on the JSSP. To investigate the influence of seeding on the optimisation, a varying percentage of seeding was used, ranging from 25 to 75\%. Interestingly, it was also pointed out that a 100\% seed is not necessarily very successful on either problems~\cite{keedwell2005hybrid}. This is one of the very few reported that seeding can in some cases beneficial to the optimisation process, but not necessarily always is. 
In~\cite{hatzakis2006} a seeding technique for dynamic environments was investigated. There, the population was seeded when a change in the objective landscape arrived, aiming at a faster convergence to the new global optimum. Again, some of the investigated seeding approaches were more successful than others.

One of the very few studies that can be found on seeding techniques for MOEAs is the one performed by \citet{hernandez2008}. There, seeds were created using gradient-based information. These were then fed into NSGA-II~\cite{NSGA-II} and the quality was assessed on the benchmark family ZDT~\cite{ZDT}. The results indicate that the proposed approach can produce a significant reduction in the computational cost of the approach. 

In general, seeding is not well documented for multi-objective problems, even for real-world problems. If seeding is done, then typically the approach is outlined and used with the comment that it worked in ``preliminary experiments"---the reader is left in the dark on the design process behind the used seeding approach. 
This is quite striking as one expects that humans can construct a few solutions by hand, even if they do not represent the ranges of the objectives well.
The least that one should be able to do is to reuse existing designs, and to modify these iteratively towards extremes. Nevertheless, even this manual seeding is rarely reported. 

In this paper, we are going to investigate the effects of two structurally different seeding techniques for five algorithms on 48~multi-objective optimization (MOO) problems.


\subsubsection*{\bf Seeding}
As seeding we use the weighted-sum method, where
the trade-off preferences are specified by non-negative weights for each objective.
Solutions to these weighted-sums of objectives can be found with an arbitrary classical single-objective evolutionary algorithm.
In our experiments we use the CMA-ES~\cite{hansen2006eda}.
Details of the two studied weighting schemes are presented in Section~\ref{sec:seeding}.

\subsubsection*{\bf Quality measure}
There are different ways to measure the quality of the solutions. A recently very popular measure is
the hypervolume indicator, which measures the volume of the objective space dominated by the set of solutions
relative to a reference point~\citep{ZitzlerT99}. Its disadvantage is its high computational complexity~\cite{BF08CGTA,2013GECCO}
and the arbitrary choice of the reference point.  We instead consider the mathematically well founded
\emph{approximation constant}.  In fact, it is known that the worst-case approximation obtained by
optimal hypervolume distributions is asymptotically
equivalent to the best worst-case additive approximation constant
achievable by all sets of the same size~\cite{approxjournal}.
For a rigorous definition, see Section~\ref{sec:preliminaries}.
This notion of multi-objective approximation was introduced by several
authors~\cite{Hansen1980, Evtushenko1987, Reuter80, RuheFruhwirth90,Loridan84} in the 80's
and its theoretical properties have been extensively studied~\citep{Daskalakis10,Diakonikolas2009,
PapadimitriouY00,PapadimitriouYannakakis01,Vassilvitskii05}.

\subsubsection*{\bf Algorithms}
We use the jMetal framework~\cite{DurilloNA10} and its implementation of 
NSGA-II~\cite{NSGA-II},
SPEA2~\cite{SPEA2},
SMS\nobreakdash-EMOA~\cite{EmmerichBN05}, and
IBEA~\cite{ZitzlerK04}.
Additionally to these more classical MOEAs, we also study AGE~\cite{IJCAI2011}, which
aims at directly minimizing the approximation constant and
has shown to perform very well for larger dimensions~\cite{AGEparentselection,AGE2}.
For each of these algorithms we compare their regular behavior after a certain
number of iterations with their performance when initialized with a certain seeding.

\subsubsection*{\bf Benchmark families}
We compare the aforementioned algorithms on four common families of benchmark functions.
These are
DTLZ~\cite{DTLZ}, 
LZ09~\cite{journals/tec/LiZ09}, 
WFG~\cite{conf/emo/HubandBWH05}, and
ZDT~\cite{ZDT}.
While the last three families only contain two- and three-dimensional problems, 
DTLZ can be scaled to an arbitrary number of dimensions.


\section{Preliminaries} 
\label{sec:preliminaries}

We consider minimization problems with $d$ objective functions, where $d\geq2$ holds.
Each objective function $f_i\colon S \mapsto \mathds{R}$, $1 \leq i \leq d$, maps from the considered search space $S$ into the real values.
In order to simplify the presentation we only work with the dominance relation on the objective space and mention that this relation transfers to the corresponding elements of $S$.

For two points $x=(x_1,\ldots,x_d)$ and $y=(y_1,\ldots,y_d)$, with $x,y \in \mathds{R}^d$ we define the following dominance relation:
\begin{align*}
x \preceq y &\ :\Leftrightarrow\  x_i \le y_i \text{ for all } 1\leq i\leq d,\\
x \prec y   &\ :\Leftrightarrow\  x \preceq y \text{ and } x \neq y.
\end{align*}

We assess the seeding schemes and algorithms by their achieved \emph{additive approximation}
of the (known) Pareto front.
We use the following definition.
\begin{definition}
\label{def:addapp}
For finite sets $S,T \subset \R^d$,
the additive approximation of $T$ with respect to $S$ is defined as
\[\alpha(S,T) := \max_{s \in S} \min_{t \in T} \max_{1 \le i \le d} (s_i - t_i).\]
\end{definition}

\noindent
We measure the approximation constant with respect to the known Pareto front of the test functions.
The better an algorithm approximates a Pareto front, the smaller the additive approximation value is.
Perfect approximation is achieved if the additive approximation constant becomes $0$.
However, the approximation constant achievable for a (finite) population with respect to
a continuous Pareto front (consisting of an infinite number of points) is always strictly larger than~$0$.
It depends on the fitness function what is the smallest possible approximation constant achievable
with a population of bounded size.

\subsection{Seeding}
\label{sec:seeding}

For the task of computing the seeds, we employ an evolutionary strategy (ES), because it ``self-adapts" the extent to which it perturbs decision variables when generating new solutions based on previous ones. The Covariance Matrix Adaptation based evolutionary strategy (CMA-ES)~\cite{hansen2006eda} self-adapts the covariance matrix of a multivariate normal distribution. This normal distribution is then used to sample from the multidimensional search space where each variate is a search variable. The co-variance matrix allows the algorithm to respect the correlations between the variables making it a powerful evolutionary search algorithm. 

To compute a seed, a (2,4)-CMA-ES minimizes $\sum_{i=1}^{d} a_i f_i (x)$, where the $f_i(x)$ are the objective values of the solution $x$. 
In preliminary testing, we noticed that larger population values for CMA-ES tended to result in seeds with better objective values. This came at the cost of significantly increased evaluation budgets, as the learning of the correlations takes longer. Our choice does not necessarily represent the optimal choice across all 48 benchmark functions, however, it is our take on striking a balance between (1) investing evaluations in the seeding and (2) investing evaluations in the regular multi-objective optimization. 
Note that large computational budgets for the seeding have the potential to put the unseeded approaches at a disadvantage, if the final performance assessment is not done carefully. 

The number of seeds, the coefficients used, and the budget of evaluations is determined by the seeding approaches, which we will describe in the following.

\CAC: A total of 10,000 evaluations is equally distributed over the generation of $d+1$ seeds. 
The rest of the population is generated randomly. 
For the $i$-th seed, $1\leq i \leq d$, the coefficients $a_j$ ($1 \leq j \leq d$) are set in the following way:
$$
a_i =
\begin{cases}
 10 & \text{if }i=j, \\
 1 & \text{otherwise.}
\end{cases}
$$
Thus, we prevent the seeding mechanism from treating the optimization problem in a purely single-objective way by entirely neglecting any trade-off relationships between the objectives.\footnote{If the ranges of the objective values differ significantly, then the coefficients should be adjusted accordingly.} Lastly, the $(d+1)$-th weight vector uses equal weights of 1 per objective. This way, we aim at getting a seed that is relatively central with respect to the others. 

\LC: Here a total of 100~seeds is generated, where each seed is the result of running CMA-ES for 1,000 evaluations. 
The coefficients of the linear combinations are integer values and we construct them in the following way. First, we consider all ``permutations'' of coefficients with $a_i=1$ for one coefficient and $a_{j\neq i}=0$ for all others. Then, we consider all permutations where two coefficients have the value 1, then those where three coefficients have the value 1, and so on. When all such  permutations that are based on $\{0,1 \}$ are considered, we consider all permutations based on $\{0,1,2\}$, then based on $\{0,1,3\}$, then based on $\{0,2,3\}$, and so on.

Consequently, we achieve a better distribution of points in the objective space. This comes, however, at the increased initial computational cost. Furthermore, the budget per seed is lower than in the \CAC approach, which typically results in less optimized seeds.

\NS: All solutions of the initial population are generated randomly. This is the approach that is typically used for the generation of the initial population.

\subsection{Evolutionary Multi-Objective Optimization Algorithms}

In the following, we outline the five optimization algorithms for which we will later-on investigate the benefits of seeding the initial populations.

Many approaches try to produce good approximations of the true Pareto front by incorporating different preferences. 
For example, the environmental selection in NSGA-II~\cite{NSGA-II} first ranks the individuals using non-dominated sorting. Then, in order to distinguish individuals with the same rank, the crowding distance metric is used, which prefers individuals from less crowded sections of the objective space. The metric value for each solution is computed by adding the edge lengths of the cuboids in which the solutions reside, bounded by the nearest neighbors.

SPEA2~\cite{SPEA2} works similarly. The raw fitness of the individuals according to Pareto dominance relations between them is calculated, and then a density measure to break the ties is used. The individuals that reside close together in the objective space
are less likely to enter the archive of best solutions. 

In contrast to these two algorithms, IBEA~\cite{ZitzlerK04} is a general framework, which uses no explicit diversity preserving mechanism. The fitness of individuals is determined solely based on the value of a predefined indicator. Typically, implementations of IBEA come with the epsilon indicator or the hypervolume indicator, where the latter measures the volume of the dominated portion of the objective space.

SMS\nobreakdash-EMOA~\cite{EmmerichBN05} is a frequently used IBEA, which uses the hypervolume indicator directly in the search process. It is a steady-state algorithm that uses non-dominated sorting as a ranking criterion, and the hypervolume as the selection criterion to discard the individual that contributes the least hypervolume to the worst-ranked front. While SMS\nobreakdash-EMOA often outperforms its competition, its runtime unfortunately increases exponentially with the number of objectives. Nevertheless, with the use of fast approximation algorithms (e.g., \cite{BF09b,bader-faster,Ishibuchi2010}), this algorithm can be applied to solve problems with many objectives as well.

Recently, approximation-guided evolution (AGE)~\cite{IJCAI2011} has been introduced, which allows to incorporate a formal notion (such as Definition~\ref{def:addapp}) of approximation into a multi-objective algorithm. 
This approach is motivated by studies in theoretical computer science studying multiplicative and additive approximations for given multi-objective optimization problems~\citep{Diakonikolas2009,Vassilvitskii05,Cheng98,Daskalakis10}. 
As the algorithm cannot have complete knowledge about the true Pareto front,
it uses the best knowledge obtained so far during the optimization process. 
It stores an archive $A$ consisting of the non-dominated objectives vectors found so far. Its 
aim is to minimize the additive approximation $\alpha(A,P)$ of the population $P$ with respect to the archive $A$.
The experimental results presented in~\cite{IJCAI2011} show that given a fixed time budget it outperforms current state-of-the-art algorithms in terms of the desired additive approximation, as well as the covered hypervolume on standard benchmark functions. 

\section{Experimental Setup}

We use the jMetal framework~\cite{DurilloNA10}, and our code for the seeding as well 
all used seeds are available online\footnote{\url{http://cs.adelaide.edu.au/~markus/publications.html}}. 
As test problems we used the 
benchmark families 
DTLZ~\cite{DTLZ},
ZDT~\cite{ZDT},
LZ09~\cite{journals/tec/LiZ09}, and
WFG~\cite{conf/emo/HubandBWH05}, 
We used the functions DTLZ 1-4, each with 30 function variables
and with $d\in\{2,4,6,8\}$ objective values/dimensions.

In order to investigate the benefits of seeding even in the long run, 
we limit the calculations of the algorithms
to a maximum of $10^6$~fitness evaluations \emph{and}
a maximum computation time of four~hours per run.
Note that the time restriction had to be used as the runtime of some algorithms increases exponentially with respect to the size of the objective space.

AGE uses random parent selection; in all other algorithms parents are selected via a binary tournament. 
As variation operators, the polynomial mutation and the simulated binary crossover~\cite{Agrawal94simulatedbinary} were applied, 
which are both used widely in MOEAs~\cite{NSGA-II,Gong08,SPEA2}. The distribution parameters associated with the operators were $\eta_m=20.0$ and $\eta_c=20.0$. 
The crossover operator is biased towards the creation of offspring that are close to the parents, and was applied with $p_c=0.9$. The mutation operator has a specialized explorative effect for MOO problems, and was applied with $p_m=1/($number of decision variables$)$.
Population size was set to $\mu=100$ and $\lambda=100$. Each setup was repeated 100~times. Note that these parameter settings are the default settings in the jMetal framework, and they can often be found in the literature, which makes a cross-comparison easier. To the best of our knowledge, this parameter setting does not favor any particular algorithm or put one at a disadvantage, even though individual algorithms can have differing optimal settings for individual problems.

In a real-world scenario, if an algorithm is run several times (e.g.~because of restarts), the seeding might be only calculated once.
In this case, it might make sense to compare the unseeded and seeded variant of an algorithm
with the same number of fitness evaluations. However, we observed the expected outcome
that in this case seeding is almost always beneficial. We therefore consider a more difficult scenario where
the optimization is only run once and the number of fitness function evaluations used for the seeding is deduced
from the number of fitness evaluations available for the MOEA.

As pointed out earlier, we assess the seeding schemes and algorithms using the additive approximation of the Pareto front. 
However, as it is difficult to compute the exact achieved approximation constant of a known Pareto front, we approximate it. 
For the quality assessment on the LZ, WFG and ZDT functions, we compute the achieved additive approximations  with respect to the Pareto fronts given in the jMetal package.
For the DTLZ functions, we draw one million points of the front uniformly at random. and then compute the additive approximation achieved for this set.

We also measure the hypervolume for all experiments. As the behaviors of the five algorithms differ significantly, there is no single reference point that allows for a meaningful comparison of all functions. 
However, we observe the same qualitative comparison with the hypervolume as we do with the additive approximation. Therefore, we omit all hypervolume values in this paper, because the additive approximation constant gives a much better way to compare the results for these benchmark functions, where the Pareto fronts are known in advance.

In addition to calculating the average ratio of the achieved approximation constant with and without
seeding, we also perform a non-parametric test on the significance of the observed behavior.
For this, we compare the final approximation of the 100~runs without seeding and the 100~runs
with seeding using the Wilcoxon-Mann-Whitney two-sample rank-sum test
at the 95\% confidence level.

\section{Experimental Results}


Our results are summarized in Tables~1 and~2.  They compare the approximation constant
achieved with \CAC seeding (Table~1) and \LC seeding (Table~2)
with the same number of iterations without seeding.  As the seeding itself requires a number 
of fitness function evaluations ($10^4$ for \CAC and $10^5$ for \LC), we allocate the seeded algorithms
fewer fitness function evaluations. This makes it harder for the seeded algorithms to outperform
its unseeded variant, as discussed above.

\begin{figure*}[ht]
\vspace*{-15mm}\hspace*{-32mm}
\begin{minipage}{20cm}
\begin{center}
\includegraphics{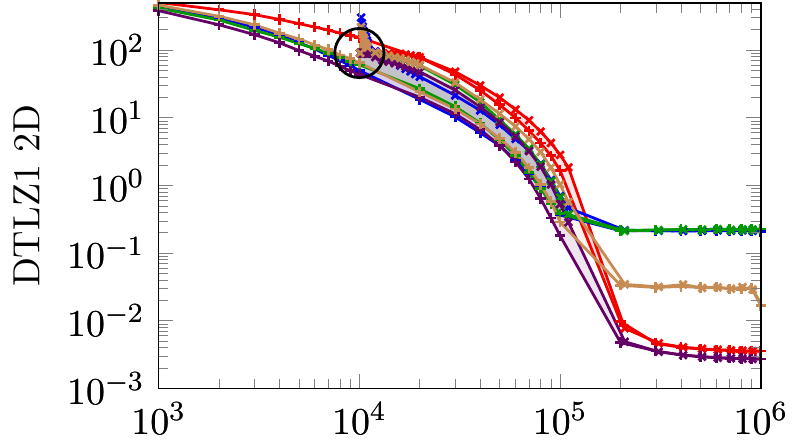}
\hspace*{.5cm}
\includegraphics{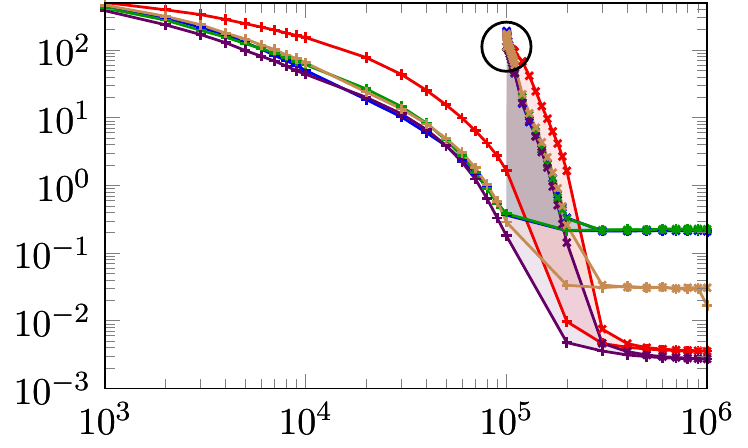}
\vspace{-5mm}
\begin{center}
\includegraphics{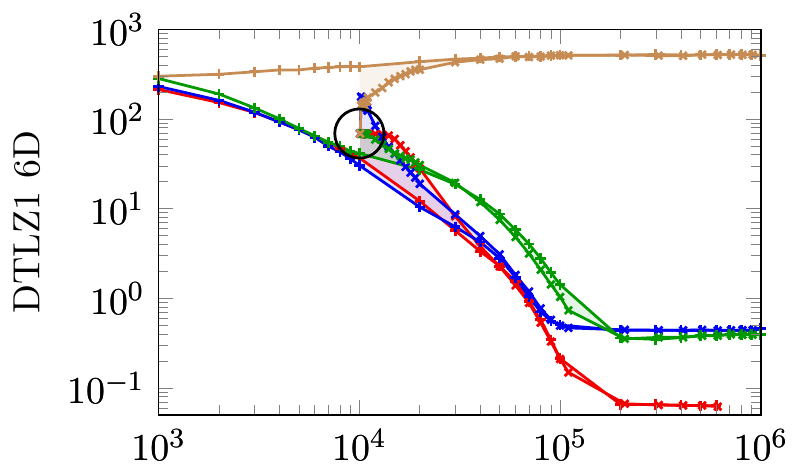}
\hspace*{.5cm}
\includegraphics{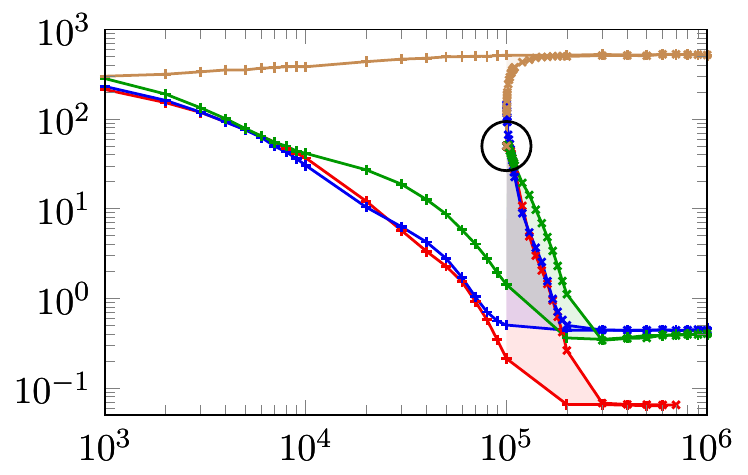}
\end{center}
\vspace{-5mm}
\begin{center}
\includegraphics{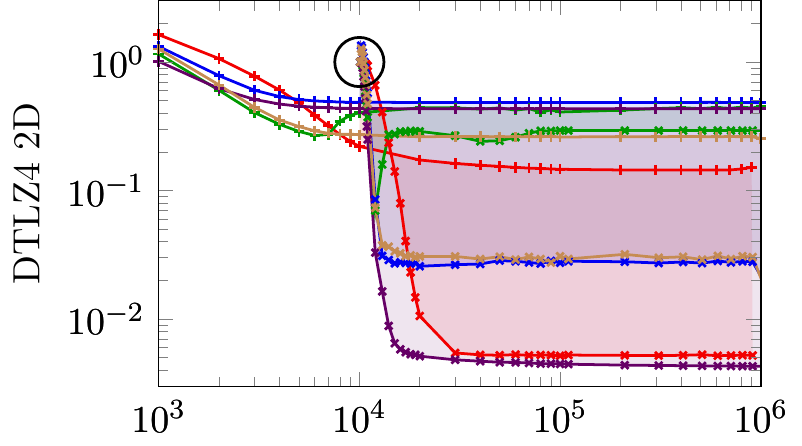}
\hspace*{.5cm}
\includegraphics{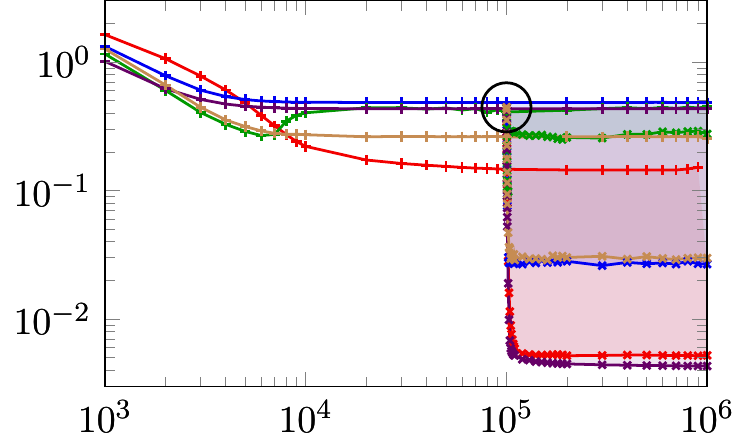}
\end{center}
\vspace{-5mm}
\begin{center}
\includegraphics{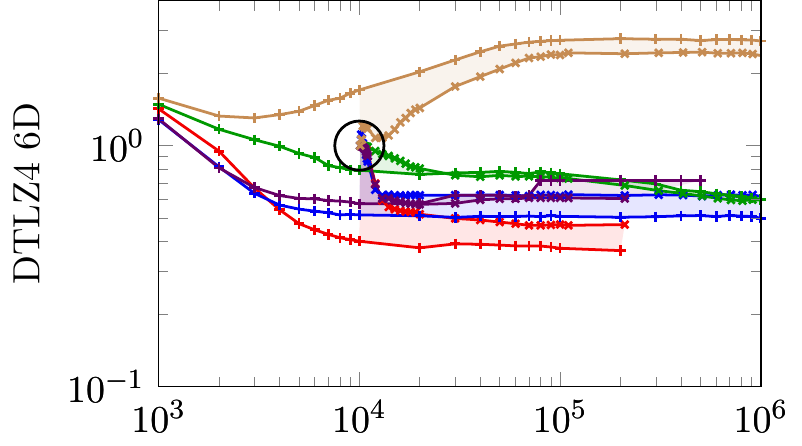}
\hspace*{.5cm}
\includegraphics{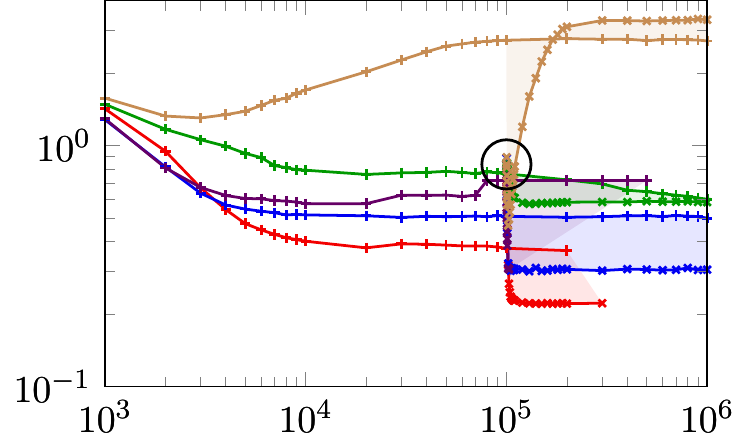}
\end{center}
\end{center}
\end{minipage}
\vspace{-0mm}
\caption[Comparison.]{Comparison of seeding with \protect\CAC (left column) and \protect\LC (right column) on DTLZ1 and DTLZ4 for two and six dimensions.
The approximation constant of the Pareto front ($y$\protect\nobreakdash-axis) is shown
as a function of the number of fitness function evaluations ($x$\protect\nobreakdash-axis)
for the seeded~($\times$) and unseeded~($+$) versions of AGE (\protect\plotAGE), IBEA (\protect\plotIBEA), NSGA-II (\protect\plotNSGATWO), SMS-EMOA (\protect\plotSMSEMOA), and SPEA2 (\protect\plotSPEATWO).
The figures show the average of 100~repetitions each.
Smaller approximation constants indicate a better approximation of the front.
The plots for the seeded versions are shifted by the number of iterations required
by the \protect\CAC seeding ($10^4$~iterations) and the \protect\LC seeding ($10^5$~iterations); circles indicate the approximation of the initial seeding.
The shaded areas illustrate the difference between seeding and no seeding for a specific algorithm.
Plots end prematurely if the time limit of four hours was reached.
}
\label{fig:DTLZ}
\end{figure*}

\begin{figure*}[ht]
\vspace*{-15mm}\hspace*{-32mm}
\begin{minipage}{20cm}
\begin{center}
\includegraphics{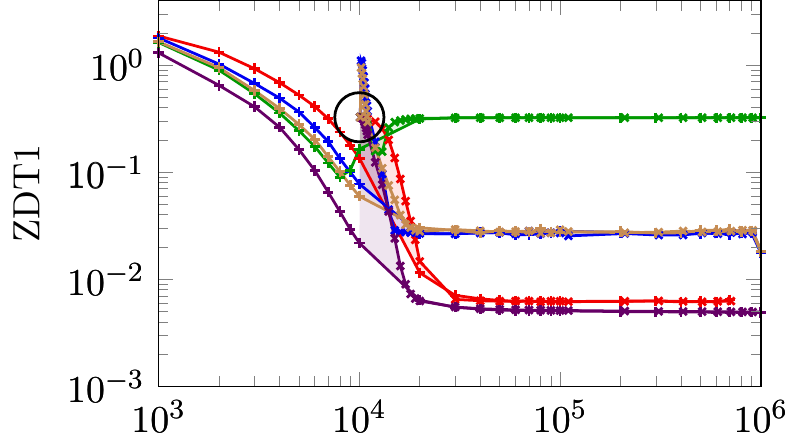}
\hspace*{.5cm}
\includegraphics{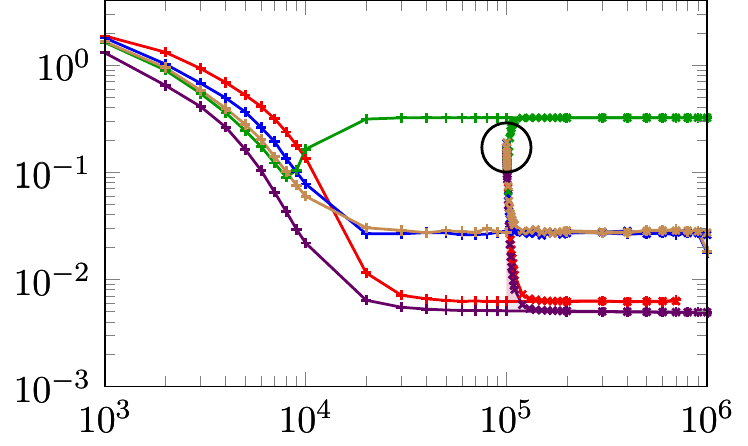}
\end{center}
\vspace{-5mm}
\begin{center}
\includegraphics{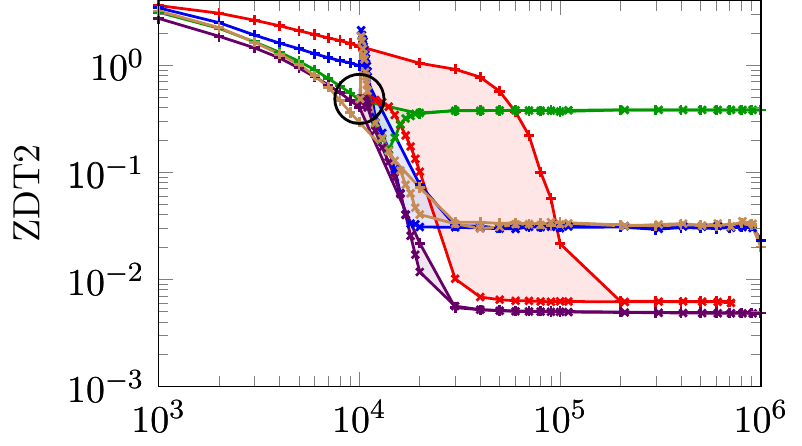}
\hspace*{.5cm}
\includegraphics{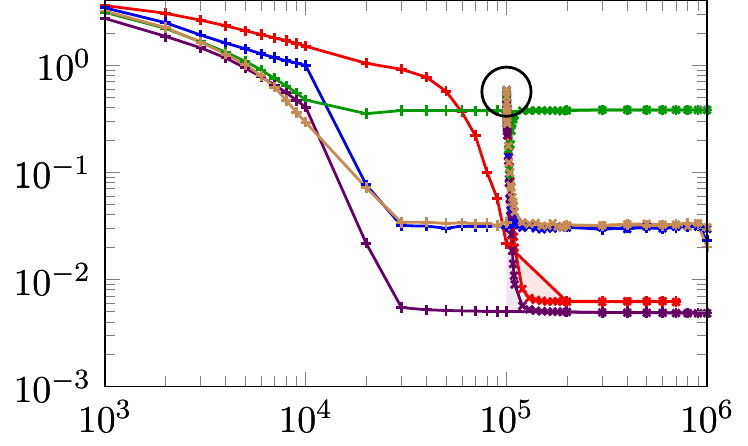}
\end{center}
\vspace{-5mm}
\begin{center}
\includegraphics{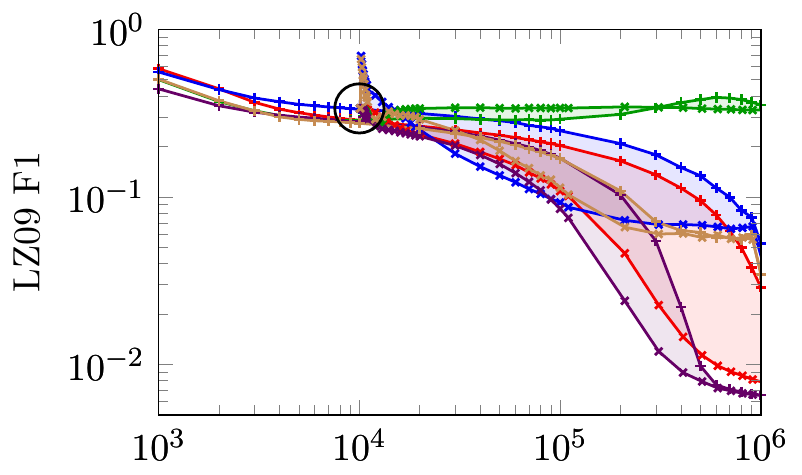}
\hspace*{.5cm}
\includegraphics{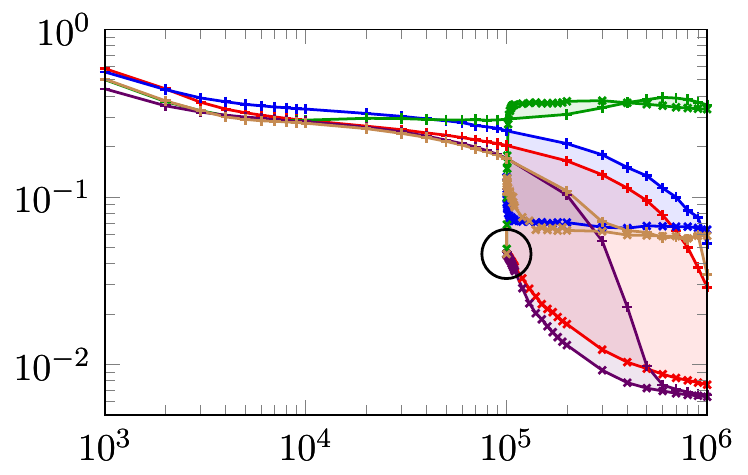}
\end{center}
\vspace{-5mm}
\begin{center}
\includegraphics{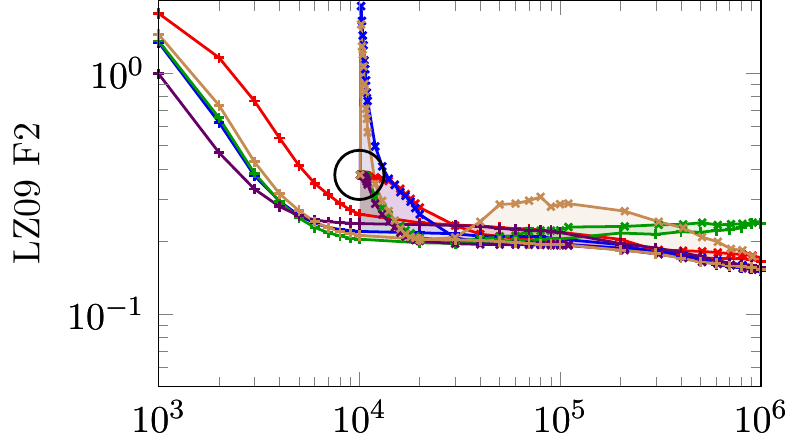}
\hspace*{.5cm}
\includegraphics{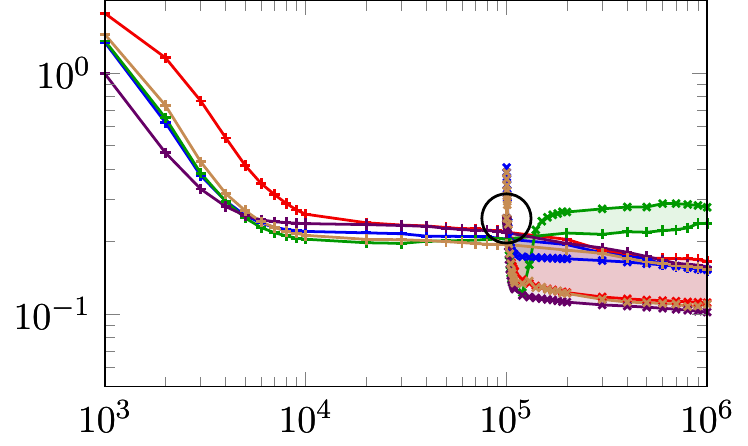}
\end{center}
\end{minipage}
\vspace{-0mm}
\caption[Comparison.]{Comparison of seeding with \protect\CAC (left column) and \protect\LC (right column) on ZDT1, ZDT2, LZ09~F1 and LZ09~F2.
The approximation constant of the Pareto front ($y$\protect\nobreakdash-axis) is shown
as a function of the number of fitness function evaluations ($x$\protect\nobreakdash-axis)
for the seeded~($\times$) and unseeded~($+$) versions of AGE (\protect\plotAGE), IBEA (\protect\plotIBEA), NSGA-II (\protect\plotNSGATWO), SMS-EMOA (\protect\plotSMSEMOA), and SPEA2 (\protect\plotSPEATWO).
The figures show the average of 100~repetitions each.
Smaller approximation constants indicate a better approximation of the front.
The plots for the seeded versions are shifted by the number of iterations required
by the \protect\CAC seeding ($10^4$~iterations) and the \protect\LC seeding ($10^5$~iterations); circles indicate the approximation of the initial seeding.
The shaded areas illustrate the difference between seeding and no seeding for a specific algorithm.
Plots end prematurely if the time limit of four hours was reached.
}
\label{fig:ZDTLZ}
\end{figure*}

Figures~\ref{fig:DTLZ} and~\ref{fig:ZDTLZ} show some representative charts
how the approximation constant behaves over the runtime of the algorithms.
First note that the approximation constant is mostly monotonically decreasing.
As a smaller approximation constant corresponds to a better approximation of the Pareto front,
this means that most algorithms achieve a better approximation over time.
Exceptions are SPEA2~(\plotSPEATWO), which is unable to handle the six dimensional variants of DTLZ,
and NSGA\nobreakdash-II (\plotNSGATWO), which sometimes gets worse after a certain time.
For most problems and algorithms, the total maximal number of fitness function evaluations ($10^6$) was enough such that
the algorithms have converged.

The small black circles in Figures~\ref{fig:DTLZ} and~\ref{fig:ZDTLZ} indicate the average approximation constant
of the initial seeding after the number of fitness function evaluations needed to calculate it. Note that because of
their specific selection schemes, some algorithms like SPEA2~(\plotSPEATWO) and NSGA\nobreakdash-II (\plotNSGATWO) sometimes
increase the approximation constant after the initial seeding (e.g.~on LZ09~F1). Another surprising effect can be observed
on LZ09~F2. There, seeding is disadvantageous to NSGA\nobreakdash-II~(\plotNSGATWO), while it is advantageous to SPEA2~(\plotSPEATWO).

For all considered test problems (not only the ones shown in Figures~\ref{fig:DTLZ} and~\ref{fig:ZDTLZ}),
either AGE~(\plotAGE) or SMS-EMOA~(\plotSMSEMOA) reach the best approximation constant.
However, for test problems with more than two or three dimensions (cf.~Figure~\ref{fig:DTLZ}),
SMS-EMOA~(\plotSMSEMOA) fails due to the high computational cost of
calculating the hypervolume. On some problems AGE~(\plotAGE) does not finish all iterations within $10^6$ steps,
but still achieves the best approximation constant (e.g.~DTLZ1~6D and DTLZ4~6D).

For all algorithms, both seedings are beneficial on some test functions.  However, the generally more performant algorithms
(AGE and SMS-EMOA) typically gain the most from both seedings.  On some functions, these algorithms
not only achieve a better approximation faster with seeding, but it seems that all
best approximations can only be achieved with seeding (e.g.~DTLZ4). On DTLZ4 2D the gap between the approximation
constant achieved by SMS-EMOA~(\plotSMSEMOA) with and without seeding is about two orders of magnitude, which is the difference between
a very good approximation of the Pareto front and basically no approximation of the Pareto front.

Tables~1 and~2 give a numerical comparison assuming that about 10\% of the fitness evaluations
are used for seeding. The shown numbers are the ratios of the median approximation constant without seeding to the median approximation constant with seeding.
Values $>\hspace{-1mm}1.00$ indicate where seeding is beneficial in the median.
We additionally show statistically significance based on the Wilcoxon-Mann-Whitney two-sample rank-sum test
at the 95\% confidence level.
``$>$'' marks statistically significant improvements,
``$<$'' marks statistically significant worsenings,
``='' marks statistically insignificant findings.
The ratios in Table~1 correspond
to the approximation constants after $10^5$ function evaluations in the left column of Figures~\ref{fig:DTLZ} and~\ref{fig:ZDTLZ}.
The ratios in Table~2 correspond
to the approximation constants after $10^6$ function evaluations in the right column of Figures~\ref{fig:DTLZ} and~\ref{fig:ZDTLZ}.

Counting only the statistically significant results over all functions and seedings, the tables show that the majority profits
from the seeding.  The algorithms which benefit the most are 
SPEA2~(\plotSPEATWO) with $20\times$``$>$" and $11\times$``$<$",
AGE~(\plotAGE) with $28\times$``$>$" and $21\times$``$<$'', and
IBEA~(\plotIBEA) with $22\times$``$>$" and $17\times$``$<$''.
There are significant differences depending on the test function.
The LZ09 benchmark family profits the most:
Summing up the significant results for all algorithms, there are $55\times$``$>$" and $18\times$``$<$''.
Also for DTLZ4 there are  $23\times$``$>$'' and $8\times$``$<$''.
The worst performance of the seeding is achieved on the rather difficult WFG functions:
While the \CAC seeding achieves over all algorithms $10\times$``$>$'' and $5\times$``$<$',
the \LC seeding only achieves $10\times$``$>$'' and $28\times$``$<$''.

We can do a similar analysis to assess the benefits of the investigated seeding approaches. 
We observe that over all algorithms
the \CAC seeding yields in total $51\times$``$>$'' and $36\times$``$<$', which is a bit better than
the \LC seeding which yields in total $62\times$``$>$'' and $55\times$``$<$''.
In order to answer the question whether this is statistically significant, we calculate the average rank
of with and without seeding for each of the 100~runs, each of the 48~functions, and each of the 5~algorithms.
With this combined data from all runs, functions and algorithms, the
Wilcoxon-Mann-Whitney two-sample rank-sum test shows significance at the 95\% confidence level
that both seedings improve upon no seeding.


\section{Conclusions}

Seeding can result in a significant reduction of the computational cost and the number of fitness function evaluations needed.
We observe that there is an advantage on many common real-valued fitness functions even if computing an initial seeding reduces the number
of fitness function evaluations available for the MOEA. For some functions we observe a dramatic improvement in quality and
needed runtime (e.g.,~DTLZ4 and the LZ09 family). 

For practitioners, our results show that it can be worthwhile to apply some form of seeding (especially when evaluations are expensive), but also to investigate different MOEAs as well, as they have proven to benefit differently from seeding. While we observed that seeding can be very beneficial,
our experiments could not reveal why this is the case for a particular combination
of seeding, algorithm, and function landscape.
To answer this, many parts have to be studied:
the mappings that the benchmark functions create from the search spaces into the objective spaces,
the connectedness between different local Pareto fronts,
the adequacy of using CMA-ES in the seeding procedure, and much more.
As a next step towards this goal, we propose to investigate seeding for combinatorial optimization problems.



\section*{Acknowledgements}

The research leading to these results has received funding from
the Australian Research Council (ARC) under grant agreement DP140103400
and from
the European Union Seventh Framework Programme (FP7/2007-2013) under grant 
agreement no 618091 (SAGE).

\newpage


\cleardoublepage


\pagestyle{empty}

\captionsetup{width=162mm}

\clearpage
\clearpage

\newcommand{\gettable}[1]{
\scalebox{.75}{%
\begin{tabular}{!{\vrule width 1pt}l!{\vrule width 1pt}r@{\,}lr@{\,}lr@{\,}lr@{\,}lr@{\,}l!{\vrule width 1pt}}
\noalign{\hrule height 1pt}
\bf \multirow{2}*{Function} &
\multicolumn{2}{c}{\hspace*{-1.5mm}\bf AGE\hspace*{-1.5mm}} &
\multicolumn{2}{c}{\hspace*{-1.5mm}\bf IBEA\hspace*{-1.5mm}} &
\multicolumn{2}{c}{\hspace*{-2.5mm}\bf NSGA-II\hspace*{5mm}} &
\multicolumn{2}{c}{\hspace*{-7mm}\bf SMS-EMOA\hspace*{1.5mm}} &
\multicolumn{2}{c!{\vrule width 1pt}}{\hspace{-2.5mm}\bf SPEA2\hspace{-1.5mm}} \\[-.5mm]
\bf &
\multicolumn{2}{c}{\hspace*{-.5mm}\scalebox{0.7}{(\plotAGE)}\hspace*{-.5mm}} &
\multicolumn{2}{c}{\hspace*{-.5mm}\scalebox{0.7}{(\plotIBEA)}\hspace*{-.5mm}} &
\multicolumn{2}{c}{\hspace*{-6.5mm}\scalebox{0.7}{(\plotNSGATWO)}\hspace*{2mm}} &
\multicolumn{2}{c}{\hspace*{-8.5mm}\scalebox{0.7}{(\plotSMSEMOA)}\hspace*{-.5mm}} &
\multicolumn{2}{c!{\vrule width 1pt}}{\hspace*{-.5mm}\scalebox{0.7}{(\plotSPEATWO)}\hspace*{-.5mm}} \\
\noalign{\hrule height 1pt}
\input{#1.tex}
\noalign{\hrule height 1pt}
\end{tabular}}
}

\begin{landscape}
\begin{table}[btp]    
\centering%

\scalebox{.75}{%
\begin{tabular}{!{\vrule width 1pt}l!{\vrule width 1pt}r@{\,}lr@{\,}lr@{\,}lr@{\,}lr@{\,}l!{\vrule width 1pt}}
\noalign{\hrule height 1pt}
\bf \multirow{2}*{Function} &
\multicolumn{2}{c}{\hspace*{-1.5mm}\bf AGE\hspace*{-1.5mm}} &
\multicolumn{2}{c}{\hspace*{-1.5mm}\bf IBEA\hspace*{-1.5mm}} &
\multicolumn{2}{c}{\hspace*{-2.5mm}\bf NSGA-II\hspace*{5mm}} &
\multicolumn{2}{c}{\hspace*{-7mm}\bf SMS-EMOA\hspace*{1.5mm}} &
\multicolumn{2}{c!{\vrule width 1pt}}{\hspace{-2.5mm}\bf SPEA2\hspace{-1.5mm}} \\[-.5mm]
\bf &
\multicolumn{2}{c}{\hspace*{-.5mm}\scalebox{0.7}{(\plotAGE)}\hspace*{-.5mm}} &
\multicolumn{2}{c}{\hspace*{-.5mm}\scalebox{0.7}{(\plotIBEA)}\hspace*{-.5mm}} &
\multicolumn{2}{c}{\hspace*{-6.5mm}\scalebox{0.7}{(\plotNSGATWO)}\hspace*{2mm}} &
\multicolumn{2}{c}{\hspace*{-8.5mm}\scalebox{0.7}{(\plotSMSEMOA)}\hspace*{-.5mm}} &
\multicolumn{2}{c!{\vrule width 1pt}}{\hspace*{-.5mm}\scalebox{0.7}{(\plotSPEATWO)}\hspace*{-.5mm}} \\
\noalign{\hrule height 1pt}
  DTLZ1 2D    & 0.99 & $=$    & 1.00 & $=$    & 1.00 & $=$    & 1.00 & $=$    & 1.02 & $=$\\
  DTLZ1 4D    & 0.94 & $<$    & 0.98 & $=$    & 0.99 & $=$    & 1.44 & $>$    & 1.05 & $=$\\
  DTLZ1 6D    & 0.18 & $<$    & 1.00 & $=$    & 1.00 & $=$    &   -- &      & 1.01 & $=$\\
  DTLZ1 8D    & 1.02 & $=$    & 0.98 & $=$    & 1.00 & $=$    &   -- &      & 0.98 & $<$\\
\noalign{\hrule height 1pt}
  DTLZ2 2D    & 0.96 & $<$    & 1.01 & $=$    & 1.00 & $=$    & 0.99 & $<$    & 1.00 & $=$\\
  DTLZ2 4D    & 1.08 & $>$    & 1.00 & $=$    & 1.00 & $=$    & 1.01 & $>$    & 1.00 & $=$\\
  DTLZ2 6D    & 0.95 & $<$    & 1.00 & $=$    & 1.00 & $=$    &   -- &      & 1.00 & $=$\\
  DTLZ2 8D    & 0.85 & $<$    & 1.00 & $=$    & 0.98 & $=$    &   -- &      & 1.00 & $>$\\
\noalign{\hrule height 1pt}
  DTLZ3 2D    & 0.85 & $<$    & 1.00 & $=$    & 1.00 & $=$    & 1.00 & $=$    & 1.02 & $=$\\
  DTLZ3 4D    & 0.97 & $<$    & 1.00 & $=$    & 1.00 & $=$    & 0.99 & $=$    & 0.95 & $=$\\
  DTLZ3 6D    & 0.09 & $<$    & 1.00 & $=$    & 0.99 & $=$    &   -- &      & 1.00 & $=$\\
  DTLZ3 8D    & 0.86 & $<$    & 1.00 & $=$    & 1.00 & $=$    &   -- &      & 1.03 & $>$\\
\noalign{\hrule height 1pt}
  DTLZ4 2D    & 0.96 & $<$    & 1.14 & $>$    & 1.00 & $>$    & 1.02 & $>$    & 1.05 & $>$\\
  DTLZ4 4D    & 4.73 & $>$    & 3.79 & $>$    & 1.00 & $>$    & 5.24 & $>$    & 2.91 & $>$\\
  DTLZ4 6D    & 1.08 & $>$    & 2.22 & $>$    & 1.00 & $=$    & 1.91 & $>$    & 0.98 & $<$\\
  DTLZ4 8D    & 1.03 & $>$    & 1.79 & $>$    & 1.01 & $=$    &   -- &      & 1.01 & $=$\\
\noalign{\hrule height 1pt}
   LZ09 F1    & 3.56 & $>$    & 1.50 & $>$    & 1.05 & $>$    & 1.04 & $>$    & 0.97 & $=$\\
   LZ09 F2    & 1.47 & $>$    & 0.97 & $<$    & 0.73 & $<$    & 1.47 & $>$    & 1.50 & $>$\\
   LZ09 F3    & 1.06 & $>$    & 1.03 & $>$    & 0.99 & $=$    & 1.05 & $>$    & 1.01 & $=$\\
   LZ09 F4    & 6.36 & $>$    & 4.11 & $>$    & 0.96 & $<$    & 5.98 & $>$    & 6.49 & $>$\\
   LZ09 F5    & 1.13 & $>$    & 1.07 & $>$    & 0.98 & $=$    & 1.06 & $>$    & 1.04 & $>$\\
   LZ09 F6    & 1.14 & $>$    & 1.00 & $>$    & 1.00 & $=$    & 1.09 & $>$    & 1.02 & $>$\\
   LZ09 F7    & 1.52 & $>$    & 1.80 & $>$    & 0.88 & $<$    & 1.75 & $>$    & 1.58 & $>$\\
   LZ09 F8    & 0.99 & $=$    & 1.47 & $>$    & 1.09 & $=$    & 1.17 & $>$    & 0.99 & $=$\\
   LZ09 F9    & 2.16 & $>$    & 1.30 & $>$    & 0.81 & $<$    & 2.23 & $>$    & 2.22 & $>$\\

\noalign{\hrule height 1pt}
\end{tabular}}
\hspace*{5mm}

\scalebox{.75}{%
\begin{tabular}{!{\vrule width 1pt}l!{\vrule width 1pt}r@{\,}lr@{\,}lr@{\,}lr@{\,}lr@{\,}l!{\vrule width 1pt}}
\noalign{\hrule height 1pt}
\bf \multirow{2}*{Function} &
\multicolumn{2}{c}{\hspace*{-1.5mm}\bf AGE\hspace*{-1.5mm}} &
\multicolumn{2}{c}{\hspace*{-1.5mm}\bf IBEA\hspace*{-1.5mm}} &
\multicolumn{2}{c}{\hspace*{-2.5mm}\bf NSGA-II\hspace*{5mm}} &
\multicolumn{2}{c}{\hspace*{-7mm}\bf SMS-EMOA\hspace*{1.5mm}} &
\multicolumn{2}{c!{\vrule width 1pt}}{\hspace{-2.5mm}\bf SPEA2\hspace{-1.5mm}} \\[-.5mm]
\bf &
\multicolumn{2}{c}{\hspace*{-.5mm}\scalebox{0.7}{(\plotAGE)}\hspace*{-.5mm}} &
\multicolumn{2}{c}{\hspace*{-.5mm}\scalebox{0.7}{(\plotIBEA)}\hspace*{-.5mm}} &
\multicolumn{2}{c}{\hspace*{-6.5mm}\scalebox{0.7}{(\plotNSGATWO)}\hspace*{2mm}} &
\multicolumn{2}{c}{\hspace*{-8.5mm}\scalebox{0.7}{(\plotSMSEMOA)}\hspace*{-.5mm}} &
\multicolumn{2}{c!{\vrule width 1pt}}{\hspace*{-.5mm}\scalebox{0.7}{(\plotSPEATWO)}\hspace*{-.5mm}} \\
\noalign{\hrule height 1pt}
      ZDT1    & 0.99 & $<$    & 1.01 & $=$    & 1.00 & $=$    & 1.00 & $=$    & 0.98 & $=$\\
      ZDT2    & 1.00 & $=$    & 1.02 & $=$    & 1.00 & $<$    & 1.01 & $=$    & 1.02 & $=$\\
      ZDT3    & 1.02 & $=$    & 1.01 & $=$    & 1.00 & $=$    & 0.99 & $<$    & 1.00 & $=$\\
      ZDT4    & 1.01 & $=$    & 0.97 & $=$    & 1.00 & $=$    & 1.01 & $=$    & 0.99 & $=$\\
      ZDT6    & 1.01 & $=$    & 1.05 & $>$    & 1.00 & $>$    & 1.00 & $=$    & 1.01 & $=$\\
\noalign{\hrule height 1pt}
   WFG1 2D    & 1.00 & $=$    & 1.01 & $=$    & 1.00 & $=$    & 1.07 & $=$    & 1.00 & $=$\\
   WFG1 3D    & 1.01 & $=$    & 0.99 & $=$    & 1.01 & $>$    & 1.01 & $>$    & 1.01 & $=$\\
   WFG2 2D    & 0.01 & $<$    & 0.05 & $<$    & 0.47 & $<$    & 0.01 & $<$    & 0.02 & $<$\\
   WFG2 3D    & 1.00 & $=$    & 0.96 & $=$    & 1.00 & $=$    & 1.00 & $=$    & 1.02 & $=$\\
   WFG3 2D    & 1.00 & $=$    & 1.00 & $=$    & 1.00 & $=$    & 1.00 & $=$    & 1.00 & $=$\\
   WFG3 3D    & 0.98 & $=$    & 0.91 & $<$    & 1.00 & $=$    & 1.01 & $>$    & 0.97 & $=$\\
   WFG4 2D    & 0.99 & $=$    & 1.00 & $=$    & 1.00 & $=$    & 1.00 & $=$    & 0.99 & $=$\\
   WFG4 3D    & 0.99 & $=$    & 0.99 & $=$    & 1.00 & $=$    & 1.00 & $=$    & 1.01 & $=$\\
   WFG5 2D    & 1.00 & $=$    & 1.00 & $>$    & 1.00 & $>$    & 1.00 & $=$    & 1.00 & $=$\\
   WFG5 3D    & 0.98 & $=$    & 1.00 & $=$    & 1.00 & $>$    & 1.00 & $=$    & 0.98 & $=$\\
   WFG6 2D    & 0.06 & $<$    & 0.15 & $<$    & 0.73 & $<$    & 0.06 & $<$    & 0.09 & $<$\\
   WFG6 3D    & 0.38 & $<$    & 0.57 & $<$    & 0.86 & $<$    & 0.41 & $<$    & 0.63 & $<$\\
   WFG7 2D    & 1.01 & $>$    & 0.90 & $<$    & 1.00 & $>$    & 0.99 & $<$    & 1.00 & $=$\\
   WFG7 3D    & 1.00 & $=$    & 1.00 & $=$    & 1.00 & $=$    & 1.00 & $=$    & 1.00 & $=$\\
   WFG8 2D    & 0.60 & $<$    & 0.56 & $<$    & 0.96 & $<$    & 1.00 & $<$    & 1.00 & $<$\\
   WFG8 3D    & 0.87 & $<$    & 0.82 & $<$    & 1.21 & $>$    & 0.99 & $<$    & 0.98 & $<$\\
   WFG9 2D    & 1.01 & $=$    & 0.89 & $<$    & 1.00 & $=$    & 1.01 & $=$    & 1.02 & $=$\\
   WFG9 3D    & 1.01 & $>$    & 1.01 & $=$    & 1.00 & $=$    & 1.02 & $>$    & 1.00 & $=$\\

\noalign{\hrule height 1pt}
\end{tabular}}

\caption{Summary of our results for the improvement through \LC seeding.
We compare the
default strategy \NS with $10^6$ fitness function evaluations
and \LC seeding, which uses $10^5$ fitness function evaluations,
plus $9\cdot 10^5$ fitness function evaluations.
The table shows the ratio of the median approximation constant of \NS divided by the median approximation constant of \LC (100 independent runs each).
Values $>\hspace{-1mm}1.00$ indicate where \LC achieves a
better additive approximation, as the default strategy's outcome is in the dividend.
To facilitate qualitative observations, we show only two decimal place.
``$>$'' marks statistically significant improvements,
``$<$'' marks statistically significant worsenings,
``='' marks statistically insignificant findings.
In case a MOEA needed more than 4h time, the approximation constant after 4h is used. 
Dashes indicate scenarios where not even the first iteration of the algorithm
was completed within the allotted 4h.}
\label{tab:resultsLCb}
\end{table}
\end{landscape}

\begin{landscape}
\begin{table}[btp]    
\centering%

\scalebox{.75}{%
\begin{tabular}{!{\vrule width 1pt}l!{\vrule width 1pt}r@{\,}lr@{\,}lr@{\,}lr@{\,}lr@{\,}l!{\vrule width 1pt}}
\noalign{\hrule height 1pt}
\bf \multirow{2}*{Function} &
\multicolumn{2}{c}{\hspace*{-1.5mm}\bf AGE\hspace*{-1.5mm}} &
\multicolumn{2}{c}{\hspace*{-1.5mm}\bf IBEA\hspace*{-1.5mm}} &
\multicolumn{2}{c}{\hspace*{-2.5mm}\bf NSGA-II\hspace*{5mm}} &
\multicolumn{2}{c}{\hspace*{-7mm}\bf SMS-EMOA\hspace*{1.5mm}} &
\multicolumn{2}{c!{\vrule width 1pt}}{\hspace{-2.5mm}\bf SPEA2\hspace{-1.5mm}} \\[-.5mm]
\bf &
\multicolumn{2}{c}{\hspace*{-.5mm}\scalebox{0.7}{(\plotAGE)}\hspace*{-.5mm}} &
\multicolumn{2}{c}{\hspace*{-.5mm}\scalebox{0.7}{(\plotIBEA)}\hspace*{-.5mm}} &
\multicolumn{2}{c}{\hspace*{-6.5mm}\scalebox{0.7}{(\plotNSGATWO)}\hspace*{2mm}} &
\multicolumn{2}{c}{\hspace*{-8.5mm}\scalebox{0.7}{(\plotSMSEMOA)}\hspace*{-.5mm}} &
\multicolumn{2}{c!{\vrule width 1pt}}{\hspace*{-.5mm}\scalebox{0.7}{(\plotSPEATWO)}\hspace*{-.5mm}} \\
\noalign{\hrule height 1pt}
  DTLZ1 2D    & 0.99 & =    & 1.01 & =    & 1.01 & =    & 1.00 & =    & 1.01 & =\\
  DTLZ1 4D    & 1.00 & =    & 1.01 & =    & 1.00 & =    & 0.73 & $<$    & 0.96 & =\\
  DTLZ1 6D    & 4.06 & $>$    & 1.00 & =    & 1.00 & =    &   -- &      & 1.02 & =\\
  DTLZ1 8D    & 1.03 & $>$    & 0.97 & =    & 1.01 & $>$    &   -- &      & 0.99 & =\\
\noalign{\hrule height 1pt}
  DTLZ2 2D    & 1.00 & =    & 1.00 & =    & 1.00 & =    & 1.00 & =    & 1.02 & =\\
  DTLZ2 4D    & 1.06 & $>$    & 1.01 & =    & 1.00 & =    & 1.00 & =    & 1.01 & =\\
  DTLZ2 6D    & 1.05 & $>$    & 0.99 & $<$    & 0.99 & =    &   -- &      & 0.99 & =\\
  DTLZ2 8D    & 0.99 & =    & 1.00 & =    & 1.03 & =    &   -- &      & 1.01 & =\\
\noalign{\hrule height 1pt}
  DTLZ3 2D    & 0.99 & =    & 1.00 & =    & 1.00 & =    & 0.99 & =    & 1.01 & =\\
  DTLZ3 4D    & 0.99 & =    & 1.00 & =    & 1.00 & =    & 0.34 & $<$    & 0.95 & =\\
  DTLZ3 6D    & 1.27 & $>$    & 1.00 & =    & 0.99 & =    &   -- &      & 0.98 & =\\
  DTLZ3 8D    & 1.09 & $>$    & 1.00 & $<$    & 1.00 & =    &   -- &      & 1.02 & $>$\\
\noalign{\hrule height 1pt}
  DTLZ4 2D    & 0.98 & =    & 1.13 & $>$    & 1.00 & $>$    & 1.03 & $>$    & 1.05 & $>$\\
  DTLZ4 4D    & 4.51 & $>$    & 0.99 & $<$    & 0.99 & =    & 5.12 & $>$    & 1.00 & =\\
  DTLZ4 6D    & 0.47 & $<$    & 0.97 & $<$    & 1.00 & =    & 1.17 & $>$    & 1.05 & $>$\\
  DTLZ4 8D    & 0.98 & =    & 0.97 & $<$    & 1.05 & $>$    &   -- &      & 1.00 & =\\
\noalign{\hrule height 1pt}
   LZ09 F1    & 3.39 & $>$    & 1.57 & $>$    & 1.06 & $>$    & 1.00 & =    & 0.98 & =\\
   LZ09 F2    & 0.90 & $<$    & 1.03 & $>$    & 0.93 & $<$    & 1.08 & $>$    & 1.06 & $>$\\
   LZ09 F3    & 1.05 & $>$    & 0.99 & =    & 0.93 & $<$    & 1.02 & =    & 0.97 & =\\
   LZ09 F4    & 2.84 & $>$    & 2.46 & $>$    & 1.03 & =    & 2.97 & $>$    & 3.49 & $>$\\
   LZ09 F5    & 0.97 & $<$    & 0.91 & $<$    & 1.02 & =    & 0.89 & $<$    & 0.88 & $<$\\
   LZ09 F6    & 0.94 & $<$    & 1.00 & $>$    & 1.00 & =    & 0.89 & $<$    & 0.28 & $<$\\
   LZ09 F7    & 1.08 & $>$    & 0.90 & $>$    & 0.93 & $<$    & 1.29 & $>$    & 1.18 & $>$\\
   LZ09 F8    & 1.07 & $>$    & 1.69 & $>$    & 0.74 & $<$    & 1.33 & $>$    & 1.10 & $>$\\
   LZ09 F9    & 1.76 & $>$    & 1.55 & $>$    & 0.83 & $<$    & 1.68 & $>$    & 1.70 & $>$\\

\noalign{\hrule height 1pt}
\end{tabular}}
\hspace*{5mm}

\scalebox{.75}{%
\begin{tabular}{!{\vrule width 1pt}l!{\vrule width 1pt}r@{\,}lr@{\,}lr@{\,}lr@{\,}lr@{\,}l!{\vrule width 1pt}}
\noalign{\hrule height 1pt}
\bf \multirow{2}*{Function} &
\multicolumn{2}{c}{\hspace*{-1.5mm}\bf AGE\hspace*{-1.5mm}} &
\multicolumn{2}{c}{\hspace*{-1.5mm}\bf IBEA\hspace*{-1.5mm}} &
\multicolumn{2}{c}{\hspace*{-2.5mm}\bf NSGA-II\hspace*{5mm}} &
\multicolumn{2}{c}{\hspace*{-7mm}\bf SMS-EMOA\hspace*{1.5mm}} &
\multicolumn{2}{c!{\vrule width 1pt}}{\hspace{-2.5mm}\bf SPEA2\hspace{-1.5mm}} \\[-.5mm]
\bf &
\multicolumn{2}{c}{\hspace*{-.5mm}\scalebox{0.7}{(\plotAGE)}\hspace*{-.5mm}} &
\multicolumn{2}{c}{\hspace*{-.5mm}\scalebox{0.7}{(\plotIBEA)}\hspace*{-.5mm}} &
\multicolumn{2}{c}{\hspace*{-6.5mm}\scalebox{0.7}{(\plotNSGATWO)}\hspace*{2mm}} &
\multicolumn{2}{c}{\hspace*{-8.5mm}\scalebox{0.7}{(\plotSMSEMOA)}\hspace*{-.5mm}} &
\multicolumn{2}{c!{\vrule width 1pt}}{\hspace*{-.5mm}\scalebox{0.7}{(\plotSPEATWO)}\hspace*{-.5mm}} \\
\noalign{\hrule height 1pt}
      ZDT1    & 0.99 & $<$    & 1.01 & =    & 1.00 & =    & 0.99 & $<$    & 0.99 & $<$\\
      ZDT2    & 1.02 & =    & 1.02 & =    & 1.00 & =    & 1.01 & $>$    & 1.00 & =\\
      ZDT3    & 1.01 & =    & 0.99 & =    & 1.00 & =    & 0.99 & =    & 0.96 & =\\
      ZDT4    & 0.99 & =    & 0.98 & =    & 1.00 & =    & 1.01 & =    & 0.99 & =\\
      ZDT6    & 1.01 & =    & 1.03 & =    & 1.00 & $>$    & 1.00 & =    & 1.00 & =\\
\noalign{\hrule height 1pt}
   WFG1 2D    & 1.02 & $>$    & 1.01 & =    & 0.87 & =    & 1.06 & $>$    & 1.00 & =\\
   WFG1 3D    & 1.00 & =    & 0.98 & $<$    & 1.00 & =    & 1.00 & =    & 1.02 & =\\
   WFG2 2D    & 0.98 & =    & 1.00 & =    & 1.00 & =    & 1.01 & =    & 1.00 & =\\
   WFG2 3D    & 1.00 & =    & 0.93 & =    & 1.00 & =    & 1.00 & =    & 0.98 & =\\
   WFG3 2D    & 1.00 & =    & 1.00 & =    & 1.00 & =    & 1.00 & =    & 1.00 & =\\
   WFG3 3D    & 1.01 & =    & 1.00 & =    & 1.00 & =    & 1.00 & =    & 0.97 & =\\
   WFG4 2D    & 1.00 & =    & 1.00 & =    & 1.00 & =    & 1.01 & $>$    & 0.97 & $<$\\
   WFG4 3D    & 1.00 & =    & 0.98 & =    & 1.00 & =    & 1.00 & =    & 1.01 & =\\
   WFG5 2D    & 1.00 & =    & 1.02 & $>$    & 1.00 & $>$    & 1.00 & =    & 1.00 & =\\
   WFG5 3D    & 0.99 & =    & 1.00 & =    & 1.00 & $>$    & 1.00 & =    & 0.98 & =\\
   WFG6 2D    & 0.93 & =    & 0.95 & =    & 1.00 & =    & 0.90 & =    & 0.96 & =\\
   WFG6 3D    & 1.01 & =    & 0.99 & =    & 1.00 & $<$    & 1.00 & =    & 0.99 & =\\
   WFG7 2D    & 1.01 & =    & 1.00 & =    & 1.00 & =    & 0.99 & $<$    & 1.01 & =\\
   WFG7 3D    & 1.00 & =    & 1.02 & $>$    & 1.00 & =    & 1.00 & =    & 1.03 & $>$\\
   WFG8 2D    & 1.02 & =    & 0.70 & $<$    & 0.96 & $>$    & 1.00 & =    & 1.79 & =\\
   WFG8 3D    & 1.18 & =    & 1.00 & =    & 1.21 & $>$    & 1.00 & =    & 1.26 & $>$\\
   WFG9 2D    & 1.00 & =    & 1.03 & =    & 1.00 & =    & 0.99 & =    & 0.99 & =\\
   WFG9 3D    & 1.00 & =    & 1.01 & =    & 1.00 & =    & 1.00 & =    & 1.02 & =\\

\noalign{\hrule height 1pt}
\end{tabular}}

\caption{Summary of our results for the improvement through \CAC seeding.
We compare the
default strategy \NS with $10^6$ fitness function evaluations
and \CAC seeding, which uses $10^4$ fitness function evaluations,
plus $9\cdot 10^5$ fitness function evaluations.
The table shows the ratio of the median approximation constant of \NS divided by the median approximation constant of \CAC (100 runs independent each).
Values $>\hspace{-1mm}1.00$ indicate where \CAC achieves a
better additive approximation, as the default strategy's outcome is in the dividend.
To facilitate qualitative observations, we show only two decimal place.
``$>$'' marks statistically significant improvements,
``$<$'' marks statistically significant worsenings,
``='' marks statistically insignificant findings.
In case a MOEA needed more than 4h time, the approximation constant after 4h is used. 
Dashes indicate scenarios where not even the first iteration of the algorithm
was completed within the allotted 4h.}
\label{tab:resultsCACb}
\end{table}
\end{landscape}
\end{document}